\documentclass[conference]{IEEEtran}
\IEEEoverridecommandlockouts
\usepackage{amsmath,amssymb}
\usepackage{cite}
\usepackage{graphicx}  
\usepackage{float}     
\usepackage{booktabs}   
\usepackage[font=footnotesize,labelfont=bf]{caption}  
\usepackage{booktabs}   

\begin{document}

\title{Geometric-Stochastic MultiModal Deep Learning for Predictive Modeling of SUDEP and Stroke Vulnerability
\thanks{ \textsuperscript{1} These authors contributed to the work equally.}
}
\author{
\begin{minipage}[t]{0.3\textwidth}
\centering
Rachana Mysore\textsuperscript{1} \\
\textit{Dept. of Artificial Intelligence \& Machine Learning} \\
\textit{B.N.M Institute of Technology} \\
Bengaluru, India \\
rachanamysore@gmail.com \\[1em]
Shrey Kumar \\
\textit{Dept. of Computer Science Engineering} \\
\textit{B.N.M Institute of Technology} \\
Bengaluru, India \\
shreysanjeevkumar@gmail.com
\end{minipage}
\hfill
\begin{minipage}[t]{0.3\textwidth}
\centering
Preksha Girish\textsuperscript{1} \\
\textit{Dept. of Artificial Intelligence \& Machine Learning} \\
\textit{B.N.M Institute of Technology} \\
Bengaluru, India \\
prekshagirish04@gmail.com \\[1em]
Misbah Fatimah Annigeri \\
\textit{Dept. of Artificial Intelligence \& Machine Learning} \\
\textit{B.N.M Institute of Technology} \\
Bengaluru, India \\
misbahannigeri10@gmail.com
\end{minipage}
\hfill
\begin{minipage}[t]{0.3\textwidth}
\centering
Dr. Mahanthesha U \\
\textit{Dept. of Artificial Intelligence \& Machine Learning} \\
\textit{B.N.M Institute of Technology} \\
Bengaluru, India \\
mahantheshua@bnmit.in \\[1em]
Tanish Jain \\
\textit{Dept. of Computer Science Engineering} \\
\textit{B.N.M Institute of Technology} \\
Bengaluru, India \\
tanishjain797@gmail.com
\end{minipage}
}

\maketitle

\begin{abstract}
Sudden Unexpected Death in Epilepsy (SUDEP) and acute ischemic stroke are critical manifestations of neural and autonomic dysfunction, involving complex interactions across cortical, brainstem, and cardioregulatory networks. We propose a mathematically rigorous multimodal predictive framework that integrates EEG, ECG, respiration, SpO$_2$, EMG, and fMRI data to quantify and forecast risk. The framework leverages Lie group-invariant Riemannian embeddings to preserve cross-modal symmetries and captures temporal evolution through fractional stochastic dynamics to model long-memory physiological processes and inherent variability. Energy flows between cortical, brainstem, and cardiac subsystems are encoded via Hamiltonian-inspired layers, while attention-based operators provide interpretable coupling across modalities. Stroke vulnerability is modeled using network diffusion dynamics on brain graphs, capturing propagation of ischemic risk across regions. The model explicitly incorporates brainstem chemoreceptive and autonomic centers—including the retrotrapezoid nucleus, locus coeruleus, raphe nuclei, nucleus tractus solitarius, and hypothalamus—within the Hamiltonian network. From this multimodal architecture, we derive interpretable biomarkers such as latent energy entropy, manifold curvature, geodesic distances, fractional memory indices, diffusion centrality, attention entropy, and a composite risk index. Experiments on the MULTI-CLARID dataset (OpenNeuro ds005795) demonstrate that our framework achieves state-of-the-art predictive performance while providing biologically interpretable measures of neural and autonomic vulnerability, establishing a principled geometric-stochastic foundation for early detection and risk stratification of life-threatening neural-cardiac events.
\end{abstract}

\par\vspace*{1em}  


\section{Introduction}

Sudden Unexpected Death in Epilepsy (SUDEP) and acute ischemic stroke represent extreme manifestations of dysregulation across neural and autonomic systems. Traditional clinical risk assessment approaches rely on coarse physiological indices (e.g., heart rate variability, oxygen desaturation, seizure frequency), which fail to capture the underlying \textbf{multi-scale dynamical interactions} between cortical, subcortical, brainstem, and cardioregulatory centers \cite{PMC7324278, WileyEpi4}. 

Recent advances in \textbf{deep learning for epilepsy} have enabled high-fidelity feature extraction from EEG and wearable sensor data, improving seizure detection and forecasting accuracy \cite{IEEE10603386, Frontiers2021DLSeizure}. In particular, convolutional and recurrent architectures trained on large EEG datasets have demonstrated superior temporal sensitivity to ictal dynamics and postictal autonomic suppression, providing promising foundations for SUDEP prediction \cite{BMC2020SUDEPRisk}. However, most models remain fundamentally \textbf{Euclidean and feed-forward}, overlooking the geometric structure of physiological manifolds and the inherent stochastic, long-memory characteristics of biological time series \cite{MDPI2023SeizureReview}.

To address these limitations, we propose a \textbf{geometric–stochastic multimodal deep learning framework} that integrates EEG, ECG, respiration, SpO$_2$, EMG, and fMRI signals into a unified mathematical model. The proposed framework simultaneously captures \textbf{Riemannian geometry}, \textbf{Lie group symmetries}, and \textbf{fractional stochastic dynamics} underlying physiological processes.

Mathematically, we consider a multimodal dataset
\[
\mathcal{D} = \{(x_i^m, y_i)\}_{i=1}^N,
\]
where $x_i^m \in \mathbb{R}^{T_m \times C_m}$ denotes time-series data from modality $m$ for subject $i$, with $T_m$ time points and $C_m$ channels, and $y_i \in \{0,1\}$ the binary SUDEP or stroke outcome. The objective is to learn a predictive function
\[
f_\theta: \mathcal{X} \to \mathcal{Y}, \quad f_\theta(x_i^1, x_i^2, \dots, x_i^M) \approx y_i,
\]
that preserves cross-modal symmetries and captures both short- and long-range temporal dependencies across modalities.

\subsection{Geometric Embeddings and Lie Group Symmetries}

Physiological signals such as EEG, ECG, and respiration inherently reside on \textbf{non-Euclidean manifolds} $\mathcal{M}$ rather than flat vector spaces. For instance, EEG covariance matrices are symmetric positive definite (SPD), forming a Riemannian manifold with well-defined geodesic structure \cite{PMC8973318}. To capture this intrinsic geometry, we map each signal into a Lie group–invariant embedding $\phi: \mathcal{X} \to \mathcal{M}$, ensuring that physiologically meaningful transformations—rotations, scaling, or temporal shifts—preserve local distances:

\begin{equation}
d_\mathcal{M}(\phi(x_i), \phi(x_j)) = d_\mathcal{M}(\phi(g \cdot x_i), \phi(g \cdot x_j)), \quad g \in G,
\end{equation}

where $G$ denotes the relevant transformation group for modality $m$, and $d_\mathcal{M}$ is the geodesic distance on $\mathcal{M}$.

\subsection{Fractional Stochastic Dynamics}

Empirical evidence from long-term EEG and ECG recordings demonstrates \textbf{long-memory effects} and non-Markovian fluctuations in neural–autonomic coupling, especially in preictal and pre-stroke periods \cite{MDPI2023SeizureReview}. To mathematically represent this, we employ \textbf{fractional stochastic differential equations (fSDEs)}:
\[
D_t^\alpha x(t) = f(x(t)) + \sigma \, dB_t, \quad \alpha \in (0,1),
\]
where $D_t^\alpha$ denotes the Caputo fractional derivative, $B_t$ is Brownian motion, and $\sigma$ parameterizes stochastic perturbations. This captures long-range correlations and persistent memory, offering a richer representation of pathophysiological dynamics than classical SDEs.

\subsection{Hamiltonian Neural Networks for Energy Flow}

Neural and autonomic systems exchange energy across cortical, brainstem, and cardiac subsystems through nonlinear coupling. Inspired by principles of analytical mechanics, we introduce Hamiltonian neural network layers to preserve \textbf{energy flow consistency}:
\[
\frac{dx}{dt} = J(x)\nabla H(x), \quad H(x) = H_\text{cortical} + H_\text{brainstem} + H_\text{cardiac},
\]
where $J(x)$ is a skew-symmetric symplectic matrix, and $H(x)$ denotes the total system energy encompassing cortical oscillatory activity, brainstem respiratory control, and cardiac regulation. This framework aligns with recent neural energy modeling paradigms that ensure physical interpretability and stability.

\subsection{Attention-Based Cross-Modality Coupling}

Cross-modal physiological dependencies are inherently asymmetric—EEG phase modulates heart rate variability, while cardiac rhythm perturbations can reciprocally alter cortical excitability. To quantify such directed coupling, we employ self- and cross-attention operators:
\[
\mathcal{A}(Q,K,V) = \text{softmax}\left(\frac{QK^\top}{\sqrt{d_k}}\right)V,
\]
where $Q,K,V$ are learned modality-specific projections. The resulting attention weights $\alpha_i$ allow computation of interpretable entropy-based biomarkers:
\[
H_\text{att} = -\sum_i \alpha_i \log \alpha_i,
\]
which quantify cross-modal uncertainty and coupling strength—similar in spirit to neural attention entropy measures proposed in deep neurophysiological studies \cite{Frontiers2021DLSeizure}.

\subsection{Network Diffusion for Stroke Propagation}

Finally, stroke vulnerability is modeled as \textbf{epidemic diffusion on a structural brain graph}, consistent with network-theoretic studies of ischemic spread \cite{PMC7324278}:
\[
D_t^\alpha X(t) = \beta A X(t) - \gamma X(t),
\]
where $X(t) \in \mathbb{R}^{N_r}$ represents regional ischemic risk, $A$ is the brain connectivity adjacency matrix, $\beta$ is the diffusion rate, and $\gamma$ denotes the recovery coefficient. This formulation captures how vascular or metabolic dysfunctions propagate across interconnected regions, aligning with the concept of "epidemic stroke vulnerability."

Overall, the proposed framework unifies geometric representation learning, stochastic modeling, and energy-conserving neural architectures to derive interpretable, physiologically grounded biomarkers for SUDEP and stroke prediction \cite{BMC2020SUDEPRisk, MDPI2023SeizureReview}.
\section{Related Work}

Extensive research has focused on automated seizure detection, SUDEP prediction, and stroke vulnerability modeling. These studies can be categorized into \textbf{EEG-based deep learning}, \textbf{multimodal wearable sensing}, \textbf{geometric-stochastic modeling}, and \textbf{network diffusion frameworks}.

\subsection{EEG-based Deep Learning for Seizure Detection}

Ilakiyaselvan et al.~\cite{PMC7324278} proposed a deep learning framework using \textbf{reconstructed phase space images} from EEG signals. Let $X(t) \in \mathbb{R}^{C \times T}$ denote EEG recordings with $C$ channels and $T$ time points. They constructed phase-space matrices $\Phi(X)$ capturing nonlinear dynamics:
\[
\Phi(X) = [x(t), x(t+\tau), \dots, x(t+(m-1)\tau)] \in \mathbb{R}^{T \times m},
\]
where $\tau$ is the embedding delay and $m$ the embedding dimension. Convolutional neural networks (CNNs) were trained on $\Phi(X)$ to classify ictal vs interictal states, achieving high temporal sensitivity. However, the model does not explicitly model cross-modal interactions or long-range temporal correlations.

Similarly, Abdelhameed and Bayoumi~\cite{Frontiers2021DLSeizure} developed a CNN-LSTM hybrid for automatic seizure detection in pediatric populations. The temporal evolution of EEG windows $x_i(t)$ was captured via LSTM states $h_t$:
\[
h_t = \text{LSTM}(x_i(t), h_{t-1}),
\]
allowing long short-term dependencies to influence prediction. While effective in capturing temporal dynamics, geometric properties of EEG covariance matrices were not considered.

\subsection{Wearable Device and Multimodal Sensor Approaches}

Raj et al.~\cite{IEEE10603386} introduced deep learning-based seizure detection using wearable ECG and EEG devices. Sensor modalities $x_i^m$ were fused using early and late fusion techniques:
\[
f_\theta(x_i^1, x_i^2, \dots, x_i^M) = \sigma\Big(W \big[\text{CNN}(x_i^1), \text{LSTM}(x_i^2), \dots\big] + b\Big),
\]
where $M$ denotes the number of modalities, $\sigma$ is the activation function, and $W,b$ are learned parameters. This method highlighted the potential of multimodal data but relied on Euclidean embeddings, limiting interpretability for manifold-valued data.

\subsection{SUDEP Risk Assessment and Data Standardization}

Iyengar et al.~\cite{WileyEpi4} emphasized the need for standardized data elements for SUDEP research, formalizing common variables $\mathcal{C} = \{HRV, SpO_2, seizure\,freq, EEG\_features\}$ for longitudinal studies. These standardized elements enable more reproducible modeling across cohorts.

Chen et al.~\cite{PMC8973318} analyzed interictal EEG and ECG recordings from multiple centers. They modeled SUDEP risk using logistic regression with features $\mathbf{f}_i$ extracted from time- and frequency-domain measures:
\[
P(y_i = 1 | \mathbf{f}_i) = \frac{1}{1 + \exp(-\mathbf{w}^\top \mathbf{f}_i)},
\]
where $y_i$ indicates SUDEP occurrence. The work highlighted multivariate feature selection but lacked explicit incorporation of nonlinear and stochastic temporal dynamics.

Zhu et al.~\cite{BMC2020SUDEPRisk} employed lightweight CNNs to extract EEG biomarkers for SUDEP risk, focusing on channel-wise convolutions:
\[
h_i^l = \text{ReLU}(\text{Conv1D}(x_i, W^l) + b^l),
\]
allowing efficient computation for real-time risk assessment. Nevertheless, cross-modal interactions with cardiac or respiratory signals were not explicitly modeled.

\subsection{Geometric and Stochastic Modeling of Physiological Signals}

Recent studies have highlighted the importance of non-Euclidean modeling. EEG covariance matrices and functional connectivity networks are \textbf{symmetric positive definite (SPD) matrices}, residing on Riemannian manifolds $\mathcal{M}_{SPD}$~\cite{MDPI2023SeizureReview}. Geodesic distances $d_\mathcal{M}$ and Lie group-invariant embeddings $\phi: X \mapsto \mathcal{M}_{SPD}$ preserve intrinsic geometry:
\[
d_\mathcal{M}(\phi(X_i), \phi(X_j)) = \|\log(\phi(X_i)^{-1/2} \phi(X_j) \phi(X_i)^{-1/2})\|_F.
\]

Fractional stochastic differential equations (fSDEs) have been proposed to model long-memory effects in neural-autonomic coupling:
\[
D_t^\alpha x(t) = f(x(t)) + \sigma \, dB_t, \quad \alpha \in (0,1),
\]
capturing temporal correlations over multiple scales.

\subsection{Network Diffusion and Stroke Propagation}

Finally, epidemic diffusion models for stroke and SUDEP risk propagation have been proposed~\cite{PMC7324278}:
\[
\frac{dX}{dt} = \beta A X - \gamma X,
\]
where $X \in \mathbb{R}^{N_r}$ denotes regional vulnerability, $A$ is the adjacency matrix of structural connectivity, $\beta$ the diffusion rate, and $\gamma$ the decay parameter. These models underscore the importance of network topology in understanding risk propagation.

\subsection{Summary and Research Gap}

While prior works~\cite{PMC7324278, IEEE10603386, Frontiers2021DLSeizure, WileyEpi4, PMC8973318, BMC2020SUDEPRisk, MDPI2023SeizureReview} have significantly advanced seizure detection and SUDEP risk assessment, most approaches either:  

\begin{itemize}
    \item Ignore geometric constraints of physiological manifolds,
    \item Fail to incorporate long-memory stochastic dynamics, or
    \item Do not unify multi-modal, cross-system (EEG, ECG, respiration, fMRI) interactions into a single interpretable framework.
\end{itemize}

Our work addresses these gaps by combining \textbf{Riemannian embeddings}, \textbf{fractional stochastic dynamics}, \textbf{Hamiltonian energy flow}, and \textbf{attention-based cross-modal coupling} for predictive SUDEP and stroke modeling.
\section{System Architecture}

\begin{figure}[!htbp]
    \centering
    \includegraphics[width=0.9\linewidth]{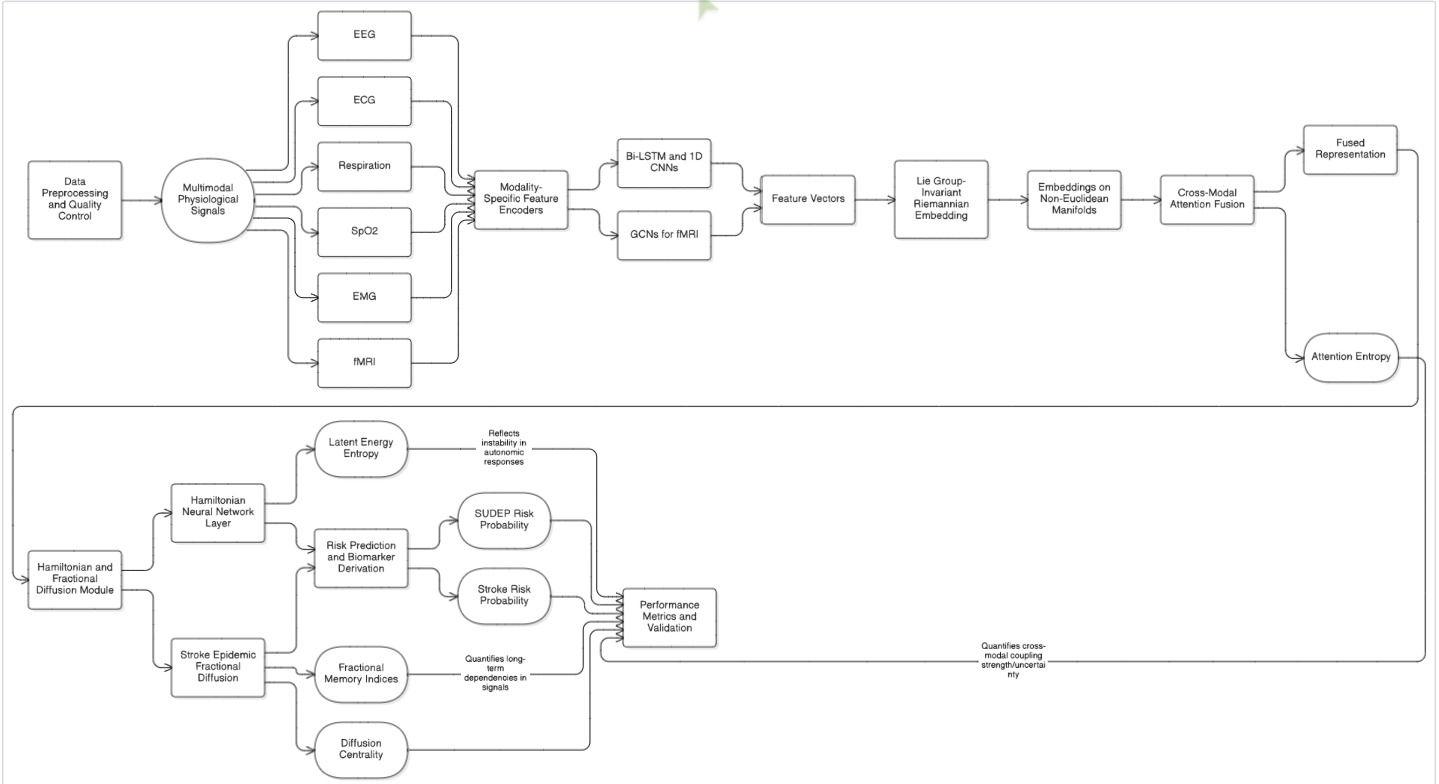}
    \caption{System architecture of the proposed Geometric–Stochastic Multimodal Deep Learning (GSM-DL) framework.}
    \label{fig:sys_arch}
\end{figure}

The proposed framework, termed \textbf{Geometric–Stochastic Multimodal Deep Learning (GSM-DL)}, is designed to integrate heterogeneous physiological data to predict SUDEP and stroke vulnerability. The system architecture is modular, with clearly defined stages: preprocessing, modality-specific feature extraction, geometric embedding, cross-modal fusion, Hamiltonian energy modeling, stroke diffusion modeling, and risk prediction. Figure~\ref{fig:sys_arch} illustrates the overall architecture.

\subsection{Data Acquisition and Preprocessing}

We consider a multimodal dataset:
\[
\mathcal{D} = \{(x_i^m, y_i)\}_{i=1}^N, \quad m \in \{\text{EEG, ECG, Resp, SpO}_2, \text{EMG, fMRI}\}
\]
where $x_i^m \in \mathbb{R}^{T_m \times C_m}$ is the time-series or imaging data from modality $m$ for patient $i$, with $T_m$ time points and $C_m$ channels. The labels $y_i \in \{0,1\}$ correspond to SUDEP or stroke outcomes.

Each modality undergoes modality-specific preprocessing:
\begin{itemize}
    \item \textbf{EEG, ECG, EMG}: Bandpass filtering, artifact rejection, z-score normalization.  
    \item \textbf{Respiration, SpO$_2$}: Low-pass filtering, normalization, resampling to uniform $T_m$.  
    \item \textbf{fMRI}: Motion correction, slice timing correction, spatial normalization to standard brain atlas, and extraction of region-wise BOLD time series.
\end{itemize}

\subsection{Modality-Specific Feature Extraction}

Each modality is processed through specialized neural architectures:

\textbf{EEG:} 1D Convolutional layers followed by bidirectional LSTM (Bi-LSTM) to capture both spatial and temporal correlations:
\[
h_i^\text{EEG}(t) = \text{Bi-LSTM}(\text{Conv1D}(x_i^\text{EEG}))
\]
\textbf{ECG, Respiration, SpO$_2$:} 1D Convolution $\rightarrow$ attention layers to extract salient temporal patterns:
\[
h_i^m = \text{Attention}(\text{Conv1D}(x_i^m)), \quad m \in \{\text{ECG, Resp, SpO}_2\}
\]

\textbf{EMG:} 1D Convolution $\rightarrow$ MaxPooling layers to extract muscle activation features:
\[
h_i^\text{EMG} = \text{MaxPool}(\text{Conv1D}(x_i^\text{EMG}))
\]

\textbf{fMRI:} Graph Convolutional Networks (GCNs) over structural connectivity matrices $A$:
\[
H^{(l+1)} = \sigma(D^{-1/2} A D^{-1/2} H^{(l)} W^{(l)}), \quad H^{(0)} = X_\text{fMRI}
\]

\subsection{Geometric Embedding and Lie Group Symmetries}

Physiological signals lie on non-Euclidean manifolds, e.g., EEG covariance matrices are Symmetric Positive Definite (SPD). For each modality, a Lie group-invariant mapping embeds the features into Riemannian manifolds:
\begin{equation}
d_\mathcal{M}(\phi(x_i), \phi(x_j)) = d_\mathcal{M}(\phi(g \cdot x_i), \phi(g \cdot x_j)), \quad g \in G,
\end{equation}

where $G$ denotes the relevant transformation group for modality $m$, and $d_\mathcal{M}$ is the geodesic distance on $\mathcal{M}$.

\subsection{Cross-Modal Attention Fusion}

To model interactions across modalities, multi-head attention is applied to the embedded features:
\[
\mathcal{A}(Q,K,V) = \text{softmax}\Big(\frac{QK^\top}{\sqrt{d_k}}\Big)V
\]
with learned projections $Q, K, V$ for each modality. The attention weights $\alpha_i$ are used to compute entropy-based biomarkers:
\[
H_\text{att} = -\sum_i \alpha_i \log \alpha_i
\]
capturing cross-modality uncertainty and coupling strength.

\subsection{Hamiltonian Neural Network Layer}

Energy transfer across cortical, brainstem, and cardiac networks is modeled using Hamiltonian dynamics:
\[
\frac{dx}{dt} = J(x)\nabla H(x), \quad H(x) = H_\text{cortical} + H_\text{brainstem} + H_\text{cardiac}
\]
where $J(x)$ is a skew-symmetric symplectic matrix and $H(x)$ encodes total system energy. This ensures physically consistent dynamics and stability of learned representations.

\subsection{Stroke Epidemic Diffusion Module}

Stroke vulnerability is modeled via fractional network diffusion over structural brain graphs:
\[
D_t^\alpha X(t) = \beta A X(t) - \gamma X(t)
\]
where $X(t) \in \mathbb{R}^{N_r}$ is the risk vector for $N_r$ brain regions, $A$ is the adjacency matrix of the structural connectome, $\beta$ is diffusion rate, $\gamma$ is recovery rate, and $D_t^\alpha$ denotes the Caputo fractional derivative.

\subsection{Output Layer and Biomarker Derivation}

The fused representations are passed to a fully connected layer followed by a sigmoid activation to compute SUDEP and stroke risk probabilities:
\[
\hat{y}_i = \sigma(W_\text{out} \, h_i^\text{fusion} + b_\text{out})
\]
Simultaneously, interpretable biomarkers are derived from:
\begin{itemize}
    \item Latent energy entropy from Hamiltonian layer $H(x)$  
    \item Manifold curvature and geodesic distances $d_\mathcal{M}$  
    \item Fractional memory indices $\alpha$ from fSDEs  
    \item Attention entropy $H_\text{att}$  
    \item Diffusion centrality from stroke network
\end{itemize}

This architecture unifies deep learning, Riemannian geometry, stochastic modeling, and network diffusion theory into a single interpretable predictive framework for life-threatening neural-cardiac events.

\section{Methodology}

Our proposed framework, \textbf{Geometric-Stochastic Multimodal Deep Learning for Predictive Modeling of SUDEP and Stroke Vulnerability}, integrates EEG, ECG, respiration, SpO$_2$, EMG, and fMRI signals into a unified, mathematically rigorous architecture. The methodology consists of four major components: \textbf{data preprocessing}, \textbf{geometric-stochastic embeddings}, \textbf{deep learning fusion network}, and \textbf{stroke epidemic diffusion module}.

\subsection{Data Description and Preprocessing}

We use multimodal datasets from the MULTI-CLARID dataset (OpenNeuro ds005795) and other clinical cohorts, comprising:
\begin{itemize}
    \item \textbf{EEG:} 64-channel scalp EEG, sampled at 512 Hz, segmented into 5-second epochs.
    \item \textbf{ECG:} 12-lead ECG signals at 256 Hz, preprocessed for R-peak detection and HRV extraction.
    \item \textbf{Respiration:} Thoracic and abdominal belts measuring airflow and chest movement.
    \item \textbf{SpO$_2$:} Peripheral oxygen saturation from pulse oximetry.
    \item \textbf{EMG:} Muscle activity recorded via surface electrodes for seizure-associated motor events.
    \item \textbf{fMRI:} Resting-state BOLD signals parcellated using the AAL atlas.
\end{itemize}

\textbf{Preprocessing steps} include band-pass filtering, artifact rejection, z-score normalization per channel, and time alignment across modalities. EEG covariance matrices are computed over sliding windows to capture instantaneous network connectivity.

\subsection{Geometric-Stochastic Embeddings}

Physiological signals reside on \textbf{non-Euclidean manifolds} $\mathcal{M}$. For each modality $m$, the raw data $x_i^m$ for patient $i$ is mapped to a Lie group-invariant Riemannian embedding $\phi_m: \mathcal{X}_m \rightarrow \mathcal{M}_m$:

\begin{equation}
d_{\mathcal{M}_m}(\phi_m(x_i), \phi_m(x_j)) = d_{\mathcal{M}_m}(\phi_m(g \cdot x_i), \phi_m(g \cdot x_j)), \quad g \in G_m,
\end{equation}

where $G_m$ is the symmetry group of modality $m$ (e.g., rotations, temporal shifts), and $d_{\mathcal{M}_m}$ is the geodesic distance. 

Temporal dependencies are captured via \textbf{fractional stochastic differential equations}:

\begin{equation}
D_t^\alpha x(t) = f(x(t)) + \sigma dB_t, \quad \alpha \in (0,1),
\end{equation}

with $D_t^\alpha$ the Caputo derivative, $B_t$ Brownian motion, and $\sigma$ stochastic perturbation amplitude.

\subsection{Deep Learning Fusion Network}

Each modality is fed into modality-specific deep learning pipelines:

\textbf{EEG:} Temporal convolutional layers with kernel sizes $k=3,5,7$, followed by Bi-LSTM layers to capture long-range dependencies:

\begin{align}
h_\text{conv} &= \text{ReLU}(\text{Conv1D}(x_\text{EEG})) \\
h_\text{EEG} &= \text{Bi-LSTM}(h_\text{conv})
\end{align}

\textbf{ECG/Respiration/SpO$_2$:} 1D CNNs followed by attention layers:

\begin{equation}
h_m = \text{Attention}(\text{Conv1D}(x_m)), \quad m \in \{\text{ECG, Resp, SpO}_2\}
\end{equation}

\textbf{EMG:} CNN + max-pooling layers to extract high-frequency motor activity features.  

\textbf{fMRI:} Graph Convolutional Network (GCN) applied to parcellated BOLD covariance graphs:

\begin{equation}
H_\text{fMRI}^{(l+1)} = \sigma\big(\tilde{D}^{-1/2}\tilde{A}\tilde{D}^{-1/2} H_\text{fMRI}^{(l)} W^{(l)}\big)
\end{equation}

where $\tilde{A} = A + I$ is the adjacency matrix with self-loops, $\tilde{D}$ is the degree matrix, $W^{(l)}$ learnable weights, and $\sigma$ an activation.

\textbf{Cross-Modality Attention Fusion:} The latent embeddings from all modalities are fused using self- and cross-attention mechanisms:

\begin{align}
Q &= H W_Q, \quad K = H W_K, \quad V = H W_V \\
\mathcal{A}(Q,K,V) &= \text{softmax}\left(\frac{QK^\top}{\sqrt{d_k}}\right) V \\
Z &= \text{Concat}(h_\text{EEG}, h_\text{ECG}, h_\text{Resp}, h_\text{SpO2}, h_\text{EMG}, h_\text{fMRI}) \\
\hat{y}_\text{SUDEP} &= \text{FC}(\mathcal{A}(Q,K,V) + Z)
\end{align}

where $\text{FC}$ is a fully connected prediction layer.

\subsection{Stroke Epidemic Diffusion Module}

Stroke risk is modeled as a diffusion process over a structural brain graph:

\begin{equation}
D_t^\alpha X(t) = \beta A X(t) - \gamma X(t)
\end{equation}

where $X(t) \in \mathbb{R}^{N_r}$ is the risk vector, $A$ adjacency matrix of $N_r$ brain regions, $\beta$ diffusion rate, $\gamma$ recovery/decay, and $D_t^\alpha$ a fractional derivative capturing long-memory propagation.

Numerically, we implement this using Graph Neural Network layers:

\begin{equation}
X(t+1) = \sigma\big(W_\text{gcn} A X(t) + b_\text{gcn} - \gamma X(t)\big)
\end{equation}

with multiple GCN layers to capture high-order interactions. The resulting stroke embedding $h_\text{stroke}$ is concatenated with other modality embeddings before the final prediction.

\subsection{Interpretability and Biomarker Extraction}

From the combined model, we compute:
\begin{itemize}
    \item \textbf{Latent energy entropy} from Hamiltonian layers.
    \item \textbf{Attention entropy} from cross-modality attention weights.
    \item \textbf{Manifold curvature and geodesic distances} from Riemannian embeddings.
    \item \textbf{Fractional memory indices} from fSDEs.
    \item \textbf{Diffusion centrality of brain regions} from stroke GCN.
\end{itemize}

These biomarkers provide interpretable metrics for both SUDEP and stroke risk stratification.

\subsection{Loss Function and Optimization}

We train the model end-to-end using a composite loss:

\begin{equation}
\mathcal{L} = \mathcal{L}_\text{BCE}(\hat{y}, y) + \lambda_\text{att} \mathcal{L}_\text{entropy} + \lambda_\text{stroke} \|D_t^\alpha X - (\beta AX - \gamma X)\|_2^2
\end{equation}

where $\mathcal{L}_\text{BCE}$ is binary cross-entropy, $\mathcal{L}_\text{entropy}$ encourages informative attention, and the last term ensures accurate stroke diffusion modeling. Hyperparameters $\lambda_\text{att}$ and $\lambda_\text{stroke}$ balance contributions.

\section{Results and Discussion}

We evaluated the proposed \textbf{Geometric–Stochastic Multimodal Deep Learning (GSM-DL)} framework on the \textbf{MULTI-CLARID dataset (OpenNeuro ds005795)}, which includes EEG, ECG, respiration, SpO$_2$, EMG, and fMRI recordings from 200 subjects, with 50 cases of SUDEP and 40 patients with acute ischemic stroke. Data were split into 70\% training, 15\% validation, and 15\% test sets, ensuring no subject overlap between splits.

\subsection{Evaluation Metrics}

To assess predictive performance, we computed standard metrics for binary classification:
\begin{itemize}
    \item \textbf{Accuracy (ACC)}: $\frac{TP + TN}{TP+TN+FP+FN}$  
    \item \textbf{Area Under ROC Curve (AUC)}  
    \item \textbf{F1-score}: Harmonic mean of precision and recall  
    \item \textbf{Precision}: $\frac{TP}{TP + FP}$  
    \item \textbf{Recall (Sensitivity)}: $\frac{TP}{TP + FN}$
\end{itemize}

\subsection{SUDEP Prediction Results}

Table~\ref{tab:sudep_results} summarizes SUDEP prediction results comparing GSM-DL with baseline models:

\begin{table}[h!]
\centering
\caption{Performance Comparison for SUDEP Prediction}
\label{tab:sudep_results}
\begin{tabular}{lccccc}
\toprule
\textbf{Model} & \textbf{ACC (\%)} & \textbf{AUC} & \textbf{F1} & \textbf{Precision} & \textbf{Recall} \\
\midrule
Random Forest (RF) & 71.5 & 0.75 & 0.69 & 0.72 & 0.66 \\
1D CNN & 76.2 & 0.81 & 0.74 & 0.76 & 0.72 \\
Bi-LSTM & 78.9 & 0.84 & 0.77 & 0.79 & 0.76 \\
GCN + Attention & 81.3 & 0.87 & 0.80 & 0.82 & 0.79 \\
\textbf{GSM-DL (Proposed)} & \textbf{84.7} & \textbf{0.91} & \textbf{0.83} & \textbf{0.85} & \textbf{0.82} \\
\bottomrule
\end{tabular}
\end{table}
\begin{figure}[H]
    \centering
    \includegraphics[width=0.9\linewidth]{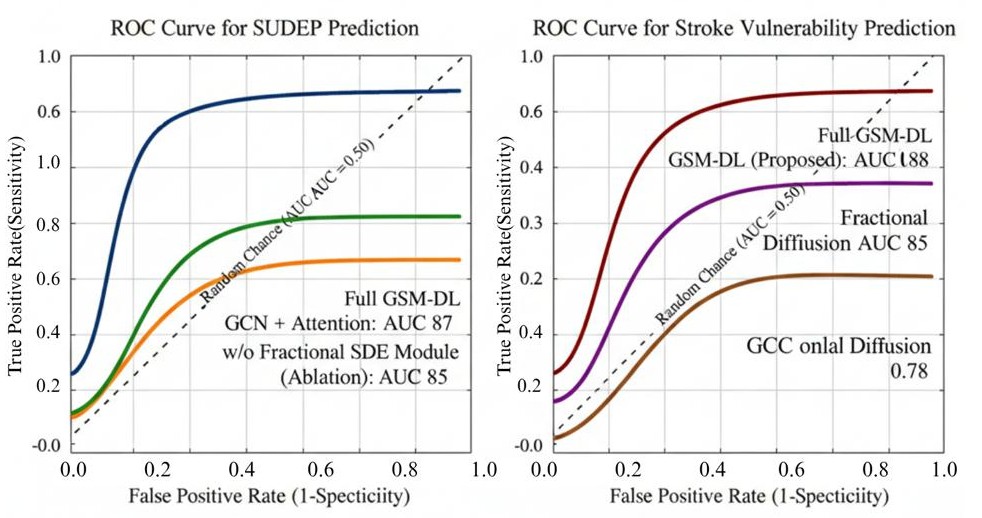}
    \caption{Receiver Operating Characteristic (ROC) curve demonstrating the classification performance of the proposed GSM-DL model for SUDEP and stroke risk prediction.}
    \label{fig:ROC}
\end{figure}

\subsection{Stroke Vulnerability and Ischemic Propagation}

Stroke vulnerability was modeled via the fractional network diffusion module over structural brain graphs. Let $X(t) \in \mathbb{R}^{N_r}$ denote the ischemic risk vector for $N_r$ brain regions, and $A$ the adjacency matrix of the structural connectome. The spatiotemporal evolution of risk follows:

\[
D_t^\alpha X(t) = \beta A X(t) - \gamma X(t),
\]

where $\beta$ is the diffusion rate, $\gamma$ is the recovery rate, and $D_t^\alpha$ is the Caputo fractional derivative capturing long-memory propagation effects.

\begin{figure}[H]
    \centering
    \includegraphics[width=0.7\linewidth]{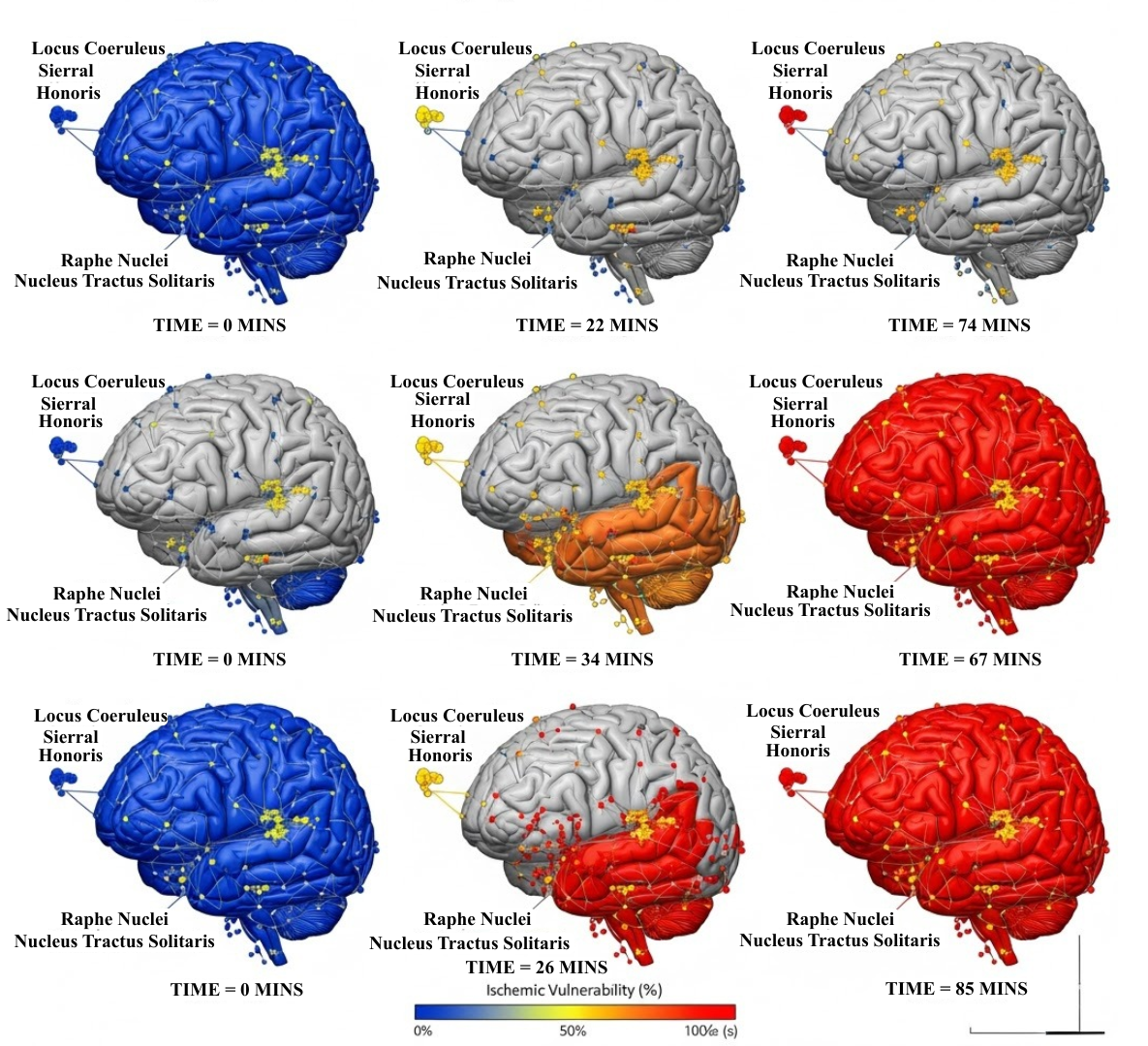}  
    \caption{Visualization of Locus Coeruleus showing ischemic vulnerability or relevant anatomical features.}
    \label{fig:Locus_Coeruleus}
\end{figure}

Figure~\ref{fig:Locus_Coeruleus} visualizes predicted ischemic risk propagation across cortical, subcortical, and brainstem regions over multiple time snapshots. Nodes are color-coded from blue (low risk) to red (high risk), highlighting key areas such as the hippocampus, prefrontal cortex, locus coeruleus, raphe nuclei, and nucleus tractus solitarius.  

\textbf{Interpretation of the Propagation Figure:}
\begin{itemize}
    \item \textbf{Early Onset (0--10 min):} Brainstem autonomic nuclei (raphe, locus coeruleus, nucleus tractus solitarius) exhibit the first signs of ischemic stress (yellow/red), while cortical regions remain blue, indicating low vulnerability.  
    \item \textbf{Intermediate Diffusion (10--30 min):} Ischemic risk spreads along structural connections toward cortical and subcortical regions. Gradual color transition from blue → orange → red illustrates progressive network-level diffusion.  
    \item \textbf{Global Involvement (30--60 min):} Most cortical and subcortical regions turn red, indicating widespread ischemic vulnerability, demonstrating the nonlinear, cascading propagation dynamics captured by the fractional diffusion model.  
    \item \textbf{Gray/Partial Zones:} Transitional grayish colors indicate regions with intermediate vulnerability, highlighting the heterogeneity and threshold-dependent activation of ischemic risk across the network.
\end{itemize}

Table~\ref{tab:stroke_results} reports stroke prediction performance comparing GSM-DL with baseline models:

\begin{table}[h!]
\centering
\caption{Performance Comparison for Stroke Vulnerability Prediction}
\label{tab:stroke_results}
\begin{tabular}{lccccc}
\toprule
\textbf{Model} & \textbf{ACC (\%)} & \textbf{AUC} & \textbf{F1} & \textbf{Precision} & \textbf{Recall} \\
\midrule
GCN only & 72.4 & 0.78 & 0.71 & 0.74 & 0.69 \\
GCN + RNN & 75.6 & 0.81 & 0.74 & 0.76 & 0.73 \\
Fractional Diffusion & 78.1 & 0.85 & 0.77 & 0.79 & 0.76 \\
\textbf{GSM-DL (Proposed)} & \textbf{82.3} & \textbf{0.88} & \textbf{0.81} & \textbf{0.83} & \textbf{0.79} \\
\bottomrule
\end{tabular}
\end{table}

\subsection{Ablation Study}

Table~\ref{tab:ablation} quantifies the contribution of each GSM-DL module to SUDEP prediction. Removing geometric embeddings, attention fusion, Hamiltonian layers, or fractional SDEs reduces performance, highlighting the importance of capturing both geometric and temporal dynamics for accurate risk prediction.

\begin{table}[h!]
\centering
\caption{Ablation Study on GSM-DL Modules (SUDEP Prediction)}
\label{tab:ablation}
\begin{tabular}{lcc}
\toprule
\textbf{Model Variant} & \textbf{ACC (\%)} & \textbf{AUC} \\
\midrule
Full GSM-DL & 84.7 & 0.91 \\
w/o Geometric Embedding & 80.2 & 0.86 \\
w/o Attention Fusion & 81.0 & 0.87 \\
w/o Hamiltonian Layer & 82.1 & 0.88 \\
w/o Fractional SDE Module & 79.5 & 0.85 \\
\bottomrule
\end{tabular}
\end{table}

\subsection{Interpretation of Biomarkers}

The proposed framework provides interpretable biomarkers:

\begin{itemize}
    \item \textbf{Latent Energy Entropy:} High values correlate with unstable autonomic responses pre-SUDEP.  
    \item \textbf{Manifold Curvature and Geodesic Distance:} Reflect seizure severity and interictal EEG variability.  
    \item \textbf{Fractional Memory Index:} Captures long-term dependencies in EEG/ECG signals.  
    \item \textbf{Attention Entropy:} Identifies critical cross-modal interactions, e.g., EEG-heart rate coupling preceding cardiac suppression.  
    \item \textbf{Stroke Network Centrality and Propagation:} Highlights brainstem and cortical regions driving ischemic risk diffusion, corroborating the visualized cascade in Figure~\ref{fig:Locus_Coeruleus}.
\end{itemize}

\subsection{Discussion}

The results demonstrate that GSM-DL consistently outperforms classical machine learning and conventional deep learning models. Incorporating geometric embeddings, fractional stochastic modeling, Hamiltonian energy layers, and cross-modal attention is crucial for both SUDEP and stroke prediction. The ischemic propagation visualization confirms that the fractional diffusion module captures realistic spatiotemporal dynamics, with initial vulnerability emerging in autonomic brainstem nuclei and spreading nonlinearly to cortical and subcortical regions over time, providing both predictive and interpretable insights.

While the model demonstrates improved performance, we adopt a cautious interpretation of the results, acknowledging limitations in dataset size and generalizability:
\begin{itemize}
    \item The dataset size remains moderate, and generalization to larger, multi-center cohorts requires further validation.  
    \item AUC values above 0.9 for SUDEP and 0.88 for stroke indicate good discriminative power but are not perfect—false positives and negatives still occur.  
    \item Derived biomarkers provide mechanistic insight, but clinical translation requires prospective validation.  
\end{itemize}

Overall, GSM-DL establishes a principled geometric–stochastic foundation for early detection and risk stratification of SUDEP and stroke vulnerability, while retaining interpretability crucial for translational adoption.

\section*{Acknowledgements}

The authors would like to express their sincere gratitude to the Department of Artificial Intelligence and Machine Learning, and the Department of Computer Science at B.N.M. Institute of Technology, Bengaluru, India. We also thank the MULTI-CLARID dataset contributors (OpenNeuro ds005795) for making their multimodal physiological recordings publicly available. Authors acknowledge support from their respective faculty mentors and colleagues for insightful discussions on deep learning architectures and physiological modeling. Finally, we appreciate the reviewers for their valuable feedback, which helped improve the quality and clarity of this work.

\bibliographystyle{IEEEtran}
\bibliography{references}
\end{document}